\documentclass[letterpaper, 10 pt, conference]{ieeeconf}  

\IEEEoverridecommandlockouts                              

\usepackage{xcolor}
\usepackage{graphicx}
\usepackage{multirow}

\usepackage{amsmath}
\usepackage{comment}
\usepackage{mathabx}
\usepackage{siunitx}
\usepackage{import}

\title{\LARGE \bf
Flow Shadowing: A Method to Detect Multiple Flow Headings using an Array of Densely Packed Whisker-inspired Sensors
}

\author{Teresa A. Kent$^{1}$
and Sarah Bergbreiter$^{2}$
\thanks{*This work was supported by MURI  award  number  FA9550-19-1-0386}
\thanks{*Corresponding Author: Teresa Kent
 {\tt\small tkent@andrew.cmu.edu}}
\thanks{$^{1}$ Robotics Institute, Carnegie Mellon University, 5000 Forbes Ave, Pittsburgh, PA, USA}
\thanks{$^{2}$ Mechanical Engineering at Carnegie Mellon University, Pittsburgh PA, USA}
}

\begin{document}

\maketitle
\thispagestyle{empty}
\pagestyle{empty}

\begin{abstract}

Understanding airflow around a drone is critical for performing advanced maneuvers while maintaining flight stability. Recent research has worked to understand this flow by employing 2D and 3D flow sensors to measure flow from a single source like wind or the drone's relative motion. Our current work advances flow detection by introducing a strategy to distinguish between two flow sources applied simultaneously from different directions. By densely packing an array of flow sensors (or whiskers), we alter the path of airflow as it moves through the array. We have named this technique ``flow shadowing'' because we take advantage of the fact that a downstream whisker shadowed (or occluded) by an upstream whisker receives less incident flow. We show that this relationship is predictable for two whiskers based on the percent of occlusion. We then show that a 2x2 spatial array of whiskers responds asymmetrically when multiple flow sources from different headings are applied to the array. This asymmetry is direction-dependent, allowing us to predict the headings of flow from two different sources, like wind and a drone's relative motion.  

\end{abstract}

\section{Introduction}

Direct measurement of airflow around a drone (e.g., the velocity and direction of flow) has led to several improvements in drone flight control \cite{simon2022flowdrone, wang2022embodied,tagliabue2020touch}. Most drones experience at least two sources of airflow: 1) the relative motion of the drone and 2) environmental sources, such as wind. Direct measurement of drone velocity through airflow can provide real-time feedback and improved state estimation \cite{tagliabue2020touch,Kim2020Whisker-inspired, fuller2014controlling}. Measurement of environmental wind flow can allow a drone to incorporate drag forces into its motion and navigation \cite{tagliabue2020touch, tagliabue2021airflow} and, importantly, improve stability \cite{paris2021autonomous, fuller2022gyroscope}.   

A measurement challenge arises when both airflow signals are present; the flow signal caused by drone motion cannot be easily distinguished from the flow signal caused by wind. No published sensor has been shown to distinguish and interpret flow from two sources simultaneously.  This gap contrasts with the natural world, where animals interpret complex flow signals during flight or while swimming using arrays of whisker/hair sensors  \cite{sterbing2011bat, boublil2021mechanosensory, camhi1969locust, zheng2023wavy}. For animals such as the harbor seal, an array of whiskers is critical to interpreting the vortices caused by prey while the seal is swimming. As a vortex passes the seal's upstream whiskers, the whiskers' shape amplifies the vortices that are then detected by the downstream whiskers \cite{zheng2023wavy}. 

\begin{figure}[!t]
\centering
\includegraphics[width=6.5 cm]{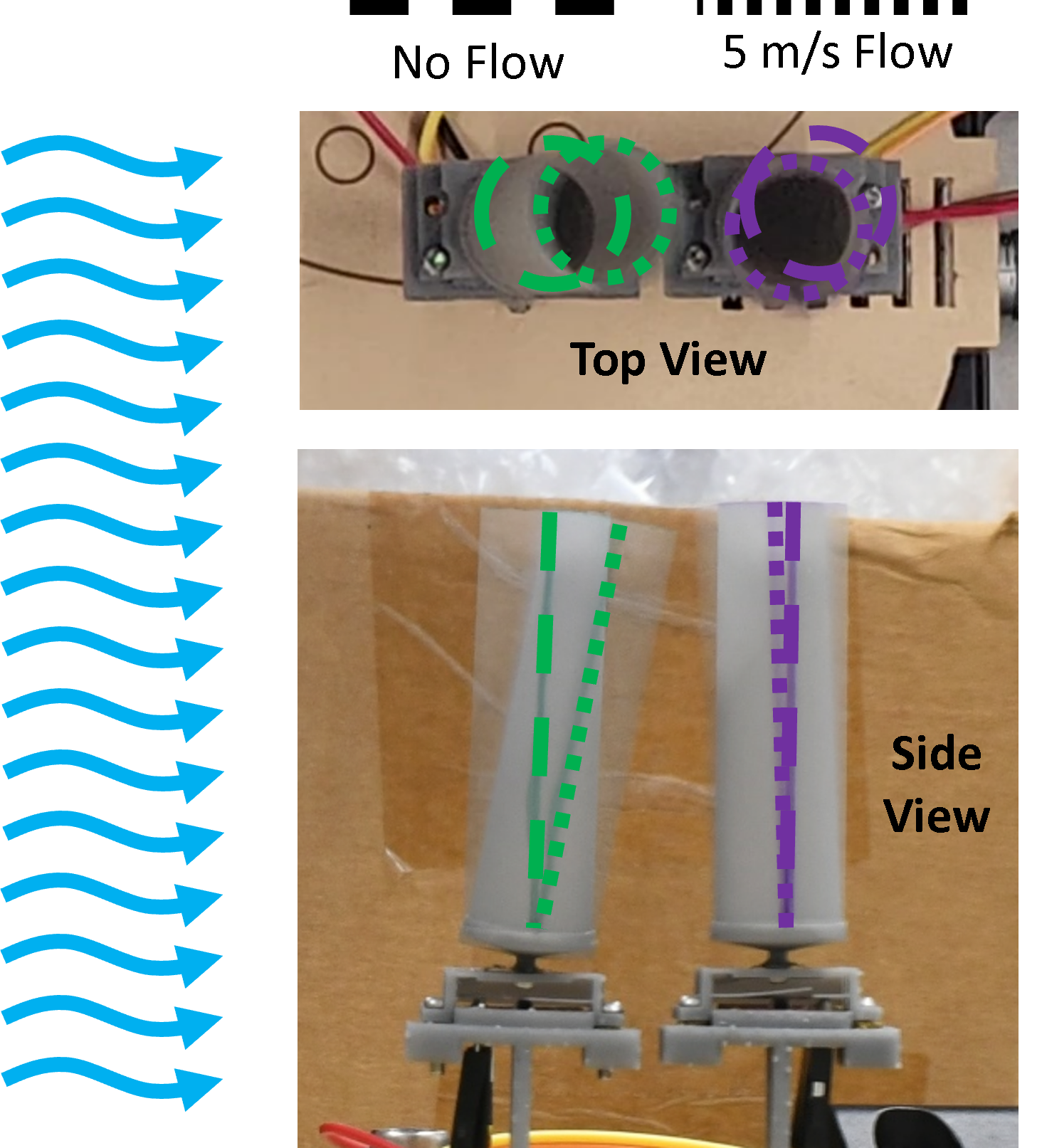}
\caption{Top and side views of two whisker-inspired flow sensors demonstrating flow shadowing. The upstream whisker (green) rotates significantly in response to an incident flow. In contrast, the downstream whisker (purple) in the shadow of the upstream whisker oscillates around its original position.}
\label{fig:Visualize Shadowing}
\end{figure}

Using the idea that the shape of an array can modify flow across an array, we engineered a sensing array with the novel ability to isolate the distinct headings of two separate flow sources. In our array, upstream whisker-inspired sensors can block or partially block the flow to downstream whiskers, a phenomenon we call ``flow shadowing''. In flow shadowing, the upstream whisker(s) receive the full effect of the airflow, but the downstream whisker(s) (which are in the front whisker's ``shadow") only receive a portion of the flow (Fig.~\ref{fig:Visualize Shadowing}). 
The shadowing becomes multi-directional when a second flow source is introduced, increasing the asymmetry in signal response. Comparing the response of all whiskers in the array allows us to identify the origin heading of two simultaneous flows.

\begin{figure*}[h]
\centering
\includegraphics[width=1.0\textwidth]{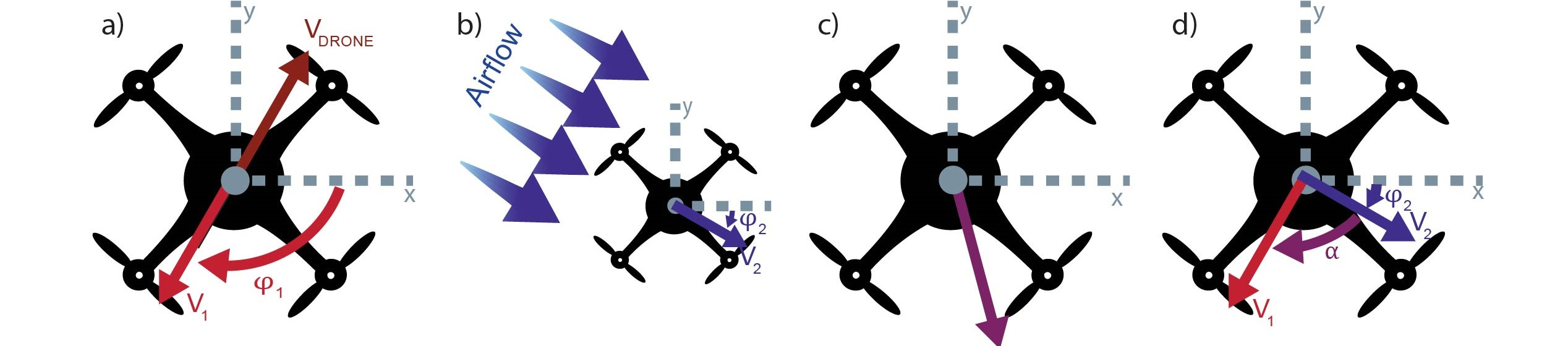}
\caption{Diagram of nomenclature for two flow signals. a) Flow 1: A drone flying at velocity $v_{Drone}$ with a flow sensor onboard measures a flow signal equal to but opposite the drone's velocity ($v_1$, $\varphi_1$). b) Flow 2: The environment in which a drone flies has wind. An onboard flow sensor will respond to the wind flow ($v_2$, $\varphi_2$). c) A flow sensor that responds to both flow types equally will measure a combined signal. d) $\alpha$ is a measure of the difference between the headings of the two flow stimuli ($\varphi_1$ or $\varphi_2$) which are estimated from an array of flow sensors. 
}
\label{fig:DroneNomenclature}
\end{figure*}

Our first experiment quantifies the flow shadowing phenomenon using a modified version of two previously developed sensors \cite{Kim2020Whisker-inspired, wang2023} under partial and complete occlusion. We found a linear response between the percent of whisker shadowed and the magnitude of the comparative signal responses. The linear relationship exists in ideal conditions (wind tunnel) and imperfect flow (box fan). Importantly, we determined that reduced spacing between the two whiskers increased the magnitude of the flow shadowing phenomenon under imperfect flow. In our second experiment, we subjected a 2x2 array of whiskers to flow from one or two different sources, showing that shadowing in multiple directions yields a predictable asymmetric response in an array. We use these asymmetric signals to estimate the flow headings under two conditions: 1) we know the heading of one of the flows, and 2) we have no prior information about the flows.

\section{Related Work}
\subsection{Arrays of hair sensors in biology}

Flow-sensing hairs are often critical to aerial flight, giving animals information about speed and stability. One study showed that when hairs were removed from bat wings, the bats altered their flight, performing more naturally stable maneuvers and flying faster \cite{sterbing2011bat}. Locusts also use hairs to detect flow and for flight stability. Airflow across locust hairs induces flight movements, which does not occur when their hairs are temporarily desensitized. Hair desensitization also prevents the locust from maintaining stable flight \cite{boublil2021mechanosensory}. 

Most animals that use whiskers/hairs to sense flow use an array of hairs. Understanding the importance of the array neurologically is complex, so much of the prior research on arrays has focused on the array mechanics. Research shows flow-sensing hairs can work together to provide a complete picture of the flow. For example, each of the flow-sensing hairs on a locust responds most strongly to flow from a single direction, but locusts are covered in hairs with different orientations, allowing them to better understand the flow environment \cite{camhi1969locust}. Harbor seals also have whisker arrays. In harbor seals, the upstream whiskers maximize the vortex-induced vibrations caused by their prey to increase the sensing ability of the downstream whiskers \cite{zheng2023wavy}. 

\subsection{Direct Flow Sensors}
There is a long history of 1D flow sensing on commercial aircraft, but here we focus on 2D and 3D flow sensing as it is more relevant to drone flight. Four types of flow sensors have been used to capture both the direction and magnitude of airflows: anemometer arrays \cite{minh20113d, bruschi2016wind,liu2014directional, arens2020measuring, piotto2011fabrication}, pitot tube pressure arrays \cite{bruschi2009low, haneda2022compact}, hot wire arrays \cite{defay2022customizable, ye2018octagon,gao2018configuration, simon2022flowdrone}, and whisker-inspired sensors \cite{Kim2020Whisker-inspired}. The first three sensors (all but the whisker-inspired sensors) use an array of 1D sensors to obtain a 2D or 3D flow vector. Our work is based on the whisker-inspired sensor; the functionality of this sensor is described in the Methods section.

Previously published flow sensors have demonstrated the benefits of drone velocity sensing \cite{Kim2020Whisker-inspired, haneda2022compact, bruschi2016wind}. The authors compared the drone velocity estimated from flow sensors to the drone's velocity profile collected from motion tracking data \cite{Kim2020Whisker-inspired, haneda2022compact}. Good velocity estimation is important for state estimation and navigation tasks. Accurate velocity estimates are also essential for advanced maneuvers, where measurements can help predict lift and drag \cite{fuller2014controlling}. All prior demonstrations of velocity estimation via flow were performed in a drone cage with negligible wind. 

Information about wind flow provides force information to a drone's flight controller. In \cite{simon2022flowdrone} and \cite{wang2022embodied}, researchers incorporated the flow sensor signal into the drone control algorithm to maintain position stability in gusty flow. In \cite{tagliabue2020touch}, researchers combined flow sensor and IMU data to estimate the direction of an external gust while subtracting flow from motion. The same researchers later showed in simulation how the flow sensor could help a drone land in gusty conditions \cite{paris2021autonomous}. In all three papers, the gust represented a significantly larger magnitude flow than any flow due to the drone's motion in contrast to the study presented in our current work.


\subsection{Indirect Measures of Flow}
While the sensors discussed in the prior section measure flow directly, drone velocity, wind \cite{tomic2022model}, and drag \cite{jia2022accurate} have also been measured through indirect measurements. The challenge, similar to direct flow measurements, is that another sensor type must be present to measure multiple flow sources \cite{neumann2015real}. Optical flow sensors estimate the velocity of a drone by tracking features on earth during the drone's flight \cite{ding2009adding, chao2014survey} provided the environment has sufficiently distinct features. Inertial measurement units (IMUs) have been used to both calculate drone velocity \cite{sani2017automatic}  and external wind \cite{neumann2015real}. Often, data from IMU sensors, GPS and optical flow \cite{bristeau2011navigation,engel2012accurate} are fused together to predict velocity, as each one is situationally imperfect at measuring velocity.

\section{Methods}

\subsection{Nomenclature}
The sensing array aims to define the origin of two flow sources of similar velocity ($v$). The headings of the two flows are labeled $\varphi_1$ and $\varphi_2$ relative to the drone. In experimental results, $\alpha$, the smallest angle between the two flows is also calculated because differentiating two separate flows becomes more difficult as $\alpha$ gets smaller. We also expect that the relative magnitudes of the two flows, $\frac{v_1}{v_2}$, will also affect our ability to accurately distinguish multiple flow sources. These variables are depicted in Fig.~\ref{fig:DroneNomenclature}.

\subsection{Array Manufacturing}
    The design of the sensing system is the same as \cite{MRLwhisker01, wang2023} and is modeled using the same methods as \cite{wang2023,MRLwhisker01}. The only update is a change in fin shape from a cross \cite{MRLwhisker01} to a \SI{50}{\milli\meter} tall, \SI{15}{\milli\meter} diameter hollow cylinder. The cylinder is expected to have a uniform drag response for various flow directions in 2D, which was not true for the cross shape \cite{tagliabue2020touch}.  
    
 A diagrammatic representation of the sensor design is in Fig. \ref{fig:ArrayDesign}c. A \SI{2}{\milli\meter^3} magnet (C0020, Supermagnetman) is glued on the backside of a laser-cut spring  (photolaser U4, LPKF) made from \SI{0.1}{\milli\meter} steel. A hollow cylinder with a taper at the base is 3D printed using a Formlabs printer and attached to the spring opposite the magnet. As the flow hits the cylinder surface, the cylinder ``whisker'' rotates about its fixed point (the center of the spring) (Supplementary Video). The magnet also rotates through an equal and opposite angle. A Hall effect sensor (TLE493-W2B6 A0, Infineon) detects this rotation. The magnet and Hall effect sensor are kept at a constant separation by a 3D-printed housing. 
    
    The array of sensors (Fig.~\ref{fig:ArrayDesign}b) maintains a spacing, $s$, between whiskers of \SI{35}{\milli\meter} unless explicitly specified otherwise.  The spacing is maintained using laser cut and 3D-printed housings and secured with screws. Each sensor in the array is connected to an Arduino Uno through a multiplexer (TCA9548A 1-to-8 I2C Multiplexer) which can sample all four sensors in the array at over 100 Hz.

    \subsection{Data Processing}
\begin{figure}[t]
\centering
\includegraphics[width=8 cm]{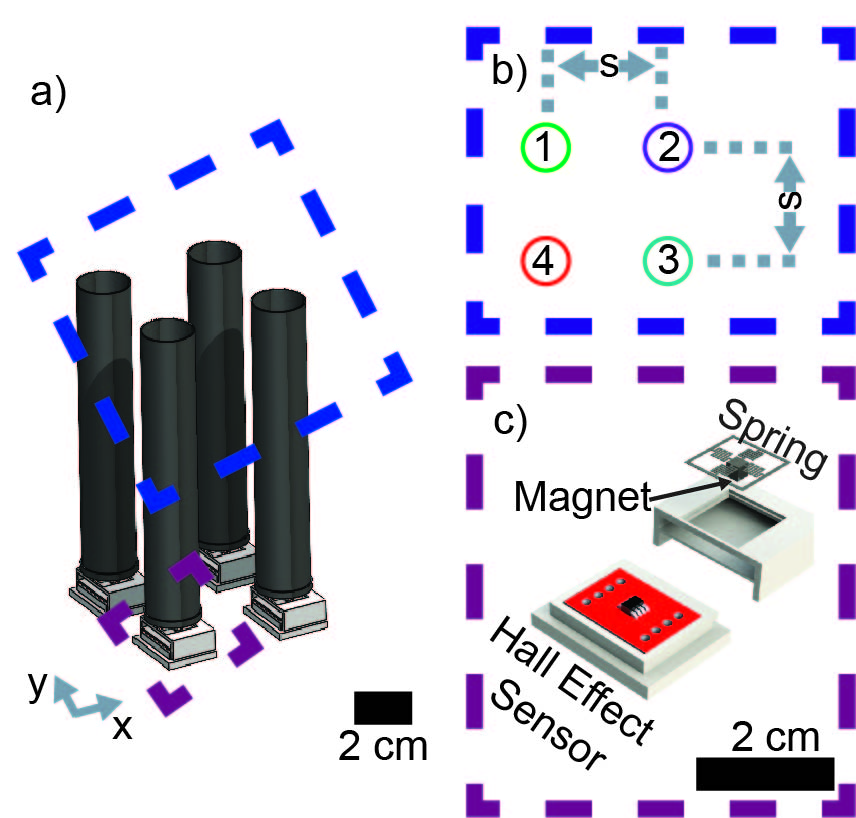}
\caption{a) 2x2 array design. b) Four whiskers are separated by $s = $ \SI{35}{\milli\meter} in both x and y axes in an array. c) The sensors are designed similar to \cite{MRLwhisker01}. A Hall effect sensor measures the rotations of the whisker drag element suspended by a spring. (Supplementary Video).}
\label{fig:ArrayDesign}
\end{figure}

    When flow hits a whisker, the whisker oscillates about a point where the spring forces and the drag forces are balanced \cite{MRLwhisker01} (Supplementary Video). The average rotation magnitude and direction are indicative of the velocity and direction of flow. These features of rotation are measured by changes in the magnetic field signal recorded by the 3 axes of the Hall effect sensor ($B_{xn}$, $B_{yn}$ and $B_{zn}$). The Hall effect sensors' axes are aligned with the axis of the full array (Fig. \ref{fig:ArrayDesign}a) and the numbers circled in Fig. \ref{fig:ArrayDesign}b represent the whisker number in the array.  
    
    To counteract variability in the manufacturing process, each sensor is calibrated individually by applying a \SI{5.5}{\meter\per\second} flow parallel to the Hall effects sensor's +x, -x, +y, and -y axes in turn. These stored values can be used  to normalize the response to analyze data. After calibrating the sensor, the magnetic field signals $B_{xn}$ and $B_{yn}$ are passed through an averaging filter with a window of 5 before solving for the magnitude of the signal ($||B_n||_2$). Subscript $n$ refers to the sensor number in the array. Averaging is important as it allows us to solve for the net vector rather than the magnitude of oscillation. Finally we solve for the direction of each sensor's signal ($\theta_n$). $\theta_n$ is always a measurement of an individual sensor and $\varphi$ (Fig.~\ref{fig:DroneNomenclature}) is always a prediction about environmental flow. 
\begin{gather}
    \theta_n=\arctan{\frac{B_{yn}}{B_{xn}}}
    \label{eqn:theta}\\
    ||B_n||_2=\sqrt{B_{xn}^2+B_{yn}^2}
    \label{eqn:BNorm}
\end{gather}
When measuring flow shadowing, the final step is to normalize the signal magnitude by $||B_{max}||_2$, the max value of $||B_n||_2$ in the array.  This step provides a magnitude relative to the array rather than relative to the calibration velocity.
    
\subsection{Experimental Setups}

\begin{figure}[h]
\centering
\includegraphics[width=8 cm]{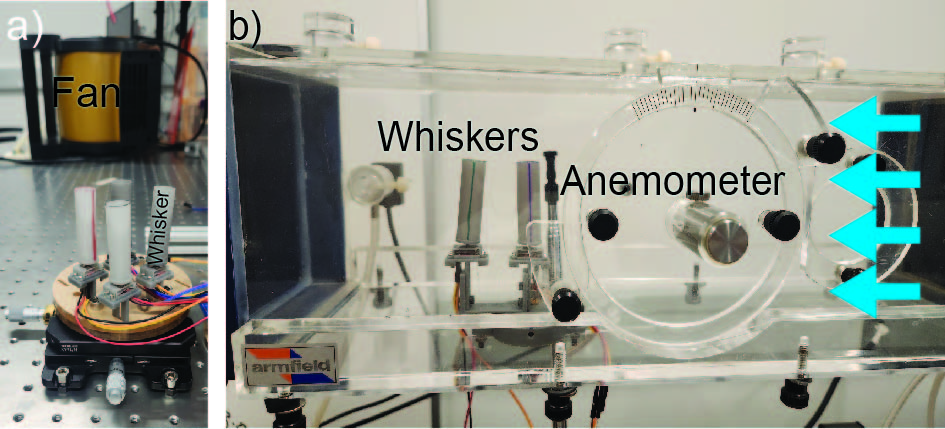}
\caption{The whisker arrays were tested under two types of airflow. a) One or two fans supplied airflow to a whisker array mounted on an optical table. The optical table and a ThorLabs rotational stage aided repeatability as the headings of flow were varied around the array. b) A wind tunnel was used to apply airflow from a single direction to whisker arrays.}
\label{fig:Experimental Setup}
\end{figure}

\subsubsection{Wind Tunnel Tests}
Wind tunnel tests (Fig.~\ref{fig:Experimental Setup}b) were carried out in a C15-10 Armfield subsonic wind tunnel. An anemometer (4330, Thomas Scientific) was placed next to the sensing array to ensure that the desired velocities were reached. The mounts for the tests were 3D printed to allow flow orientation to be controlled at intervals of \SI{5}{\degree}. Wind tunnel tests were used to evaluate the response of pairs of sensors to flow from a single direction. 

\subsubsection{Box Fan Testing}
Multiple flow directions cannot be provided in a controlled wind tunnel setup, so instead two box fans (Lasko Pro performance three-speed fan, Stanley Model 655 704 three-speed fan) were used to evaluate flow from multiple sources (Fig. \ref{fig:Experimental Setup}a). The second fan was only used to test flow from multiple sources, and we made no attempt to keep these flows laminar. Here, the array was placed on a Thorlabs manual stage which allowed for control of the incident angle of flow within \SI{0.5}{\degree} relative to the array. Fan tests were used to evaluate the response of individual sensors, pairs of sensors, and a 2x2 array of sensors to flow from one or more sources.

\subsection{Experimental Tests}
\subsubsection{Sensor Characterization}
Using the box fan setup, a single whisker was tested under single flow with orientations between \SI{0}{\degree}-\SI{360}{\degree} in increments of \SI{10}{\degree}. The test was performed at \SI{6.5}{\meter\per\second}.

\subsubsection{Flow Shadowing}
We characterized flow shadowing with pairs of whiskers using both the box fan and wind tunnel environments. Using both environments allowed us to determine how whisker sensor responses could be affected by the additional turbulence applied by the fans. Two whiskers were separated by a spacing $s =$ (\SI{35}{\milli\meter} in the wind tunnel\SI{30}{\milli\meter} using the fan). The whiskers were rotated around the upstream whisker in headings ($\varphi$) of five degree increments between \SI{-30}{\degree} and \SI{30}{\degree}. The variation in heading angle changed the percent of occlusion ($\%_{occ}$) of the downstream whisker which was calculated as follows, where $d$ represents the diameter of the drag element.
\begin{gather}
\%_{occ}=\frac{d-min\{s*sin(\varphi),d\}}{d}
\label{eqn:Oclussion}
\end{gather}

In the wind tunnel, flow velocities of both \SI{5}{\meter\per\second} and \SI{6.5}{\meter\per\second} were applied to the sensors. For the box fans, velocities of \SI{5.5}{\meter\per\second} and \SI{6.5}{\meter\per\second} were applied. First the signals were normalized compared to the values recorded during the calibration data. Second we solved for $\theta_n$ and $||B_n||_2$ using Eqns. \ref{eqn:theta},\ref{eqn:BNorm}. Finally, the value of the upstream and the downstream whiskers were compared to calculate $\frac{||B_{down}||_2}{||B_{up}||_2}$, where `up' represents the signal from the upstream whisker, e.g., the whisker closer to the fan. The tests were repeated, switching the upstream and downstream whiskers. 

\subsubsection{Spacing Tests}
We also tested how the spacing between a pair of whiskers, $s$, affected the amount of flow shadowing under 100$\%$ occlusion (Eqn.~\ref{eqn:Oclussion}, the upstream whisker completely shadowing the downstream whisker) using the fan experiment. $s$ was varied from \SI{30}{\milli\meter} to \SI{50}{\milli\meter} in \SI{5}{\milli\meter} increments (Fig.~\ref{fig:Experimental Setup}a). The applied flow speed was \SI{6.5}{\meter\per\second}. In initial tests, we found that the two whiskers made contact at this flow speed if the spacing was below \SI{30}{\milli\meter}. 

\subsubsection{Single and Double Flow Characterization}
We tested how the measured response of a 2x2 array of four whiskers (Fig.~\ref{fig:Experimental Setup}a) varied under different headings and velocities of airflow. The array was tested in all combinations of the following variables $\varphi_1=\{0^\circ,15^\circ\}$, $\alpha=\{45^\circ,90^\circ,135^\circ\}$ , $v_1=\{$ \SI{5.2}{} ,\SI{6.5}{\meter\per\second}$\}$ and $v_2=\{$\SI{7.3}{}, \SI{8.3}{\meter\per\second}\}. Combinations of the flow speeds led to $\frac{v_1}{v_2}$ ratios of $\{0.60, 0.71, 0.90\}$.  

\subsection{Algorithm}
Two algorithms were considered in this work. In Method 1, we assume that $\varphi_1$ is known (e.g., from an IMU or optic flow sensor) and we use the array results to solve for $\varphi_2$. In Method 2, we assume that no information is known about either $\varphi$ and solve for both headings. The magnitude and angle from each whisker sensor is combined into a vector $\overrightarrow{B_n}$. The two methods are outlined below, but are also visually represented in Fig. \ref{fig:Algorithm}.

\textbf{Method 1: $\varphi_1$ is known}
\begin{enumerate}
    \item Using Eqns. \ref{eqn:Oclussion}, \ref{eqn:ExpSigResponse} predict the expected response of Flow 1, $\overrightarrow{B'_n}$, with known $\varphi_1$.
    \item $\Delta\overrightarrow{B_n} = \overrightarrow{B_n}-\overrightarrow{B'_n}$
    \item Remove whisker data where $\%_{occ}=0$. 
    \item Decompose $\Delta\overrightarrow{B_n}$ into $\Delta B_y$ and $\Delta B_x$.  
    \item predicted $\varphi_2=\arctan(\frac{\sum_{i=1}^n \Delta B_y}{\sum_{i=1}^n \Delta B_x}$)
\end{enumerate}

\textbf{Method 2: Solving for both $\varphi$ values}
\begin{enumerate}
    \item Estimate $\varphi_1$: A first estimate for $\varphi_1$ can be found from the $\theta_n$ value furthest from the mean of all $\theta_n$, $\hat{\theta}$. Picking this value decreases the chance of predicting $\varphi_1$ as the net flow.
    \item Using Eqns. \ref{eqn:Oclussion}, \ref{eqn:ExpSigResponse} predict the expected response of first flow, $\overrightarrow{B'_n}$, with predicted $\varphi_1$.
    \item $\Delta\overrightarrow{B_n} = \overrightarrow{B_n}-\overrightarrow{B'_n}$
    \item Decompose $\Delta\overrightarrow{B_n}$ into $\Delta B_y$ and $\Delta B_x$
    \item predicted $\varphi_2=\arctan(\frac{\sum_{i=1}^n \Delta B_y}{\sum_{i=1}^n \Delta B_x}$)
    \item Repeat steps 2-5, starting with $\varphi_2$ as the input flow to improve upon $\varphi_1$ prediction. 
\end{enumerate}

\section{Results and Discussion}
\subsection{Single Sensor Characterization}
\begin{figure}[h]
\centering
\includegraphics[width=9
cm]{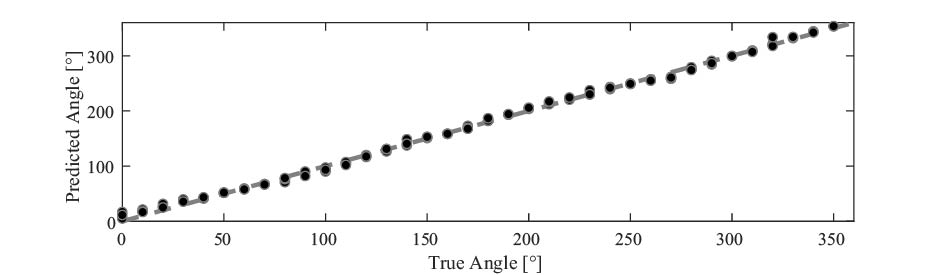}
\caption{Comparison of the true heading of flow ($\varphi$) versus the sensor predicted flow heading ($\theta$) for a single sensor with airflow provided by a fan. Each heading includes two trials.}
\label{fig:thetaAccuracy}
\end{figure}

The sensor was updated from \cite{MRLwhisker01} to obtain a uniform response from all flow headings by changing the previous cross shape to a circle. The accuracy of the sensor's $\theta$ reading was compared to the applied flow $\varphi$ and found to be accurate with a root mean square error of \SI{5.22}{\degree} (Fig.~\ref{fig:thetaAccuracy}).

\begin{figure}[h]
\centering
\includegraphics[width=8 cm]{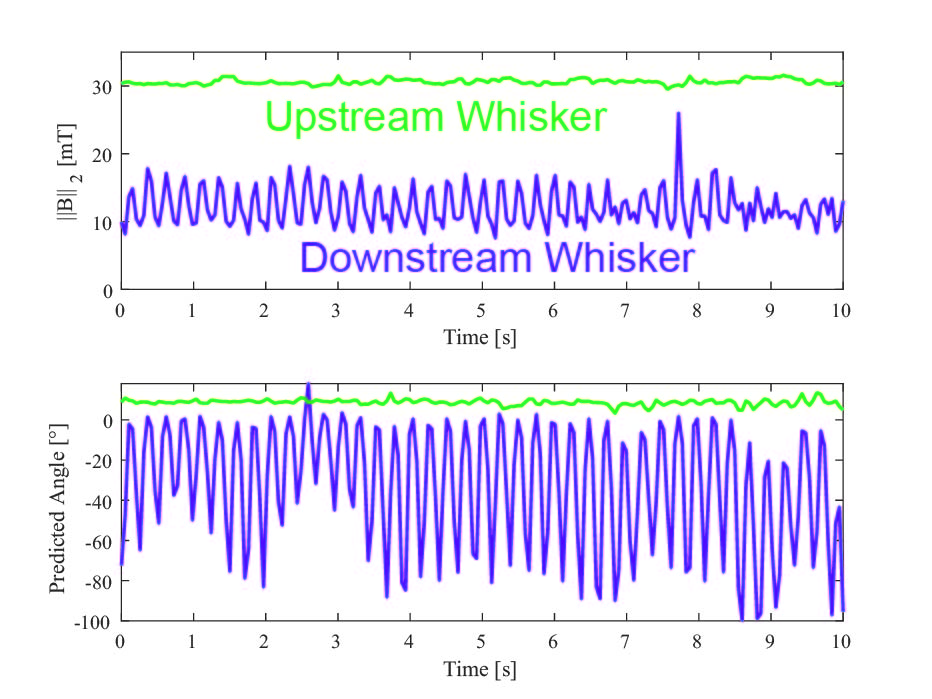}
\caption{Results from a single trial of two whiskers placed under a flow of $v=$\SI{6.5}{\meter\per\second}, $\varphi= 0^\circ$ in the wind tunnel. In this case, $\%_{occ} = $ 100 $\%$. a) Magnitude of the upstream (green) and downstream (purple) whisker response ($||B||_2$, Eqn. \ref{eqn:BNorm}). b) The estimated direction of the whiskers' responses ($\theta$, Eqn. \ref{eqn:theta}).
}
\label{fig:VisFlowShadow}
\end{figure}

\subsection{Flow Shadowing Characterization}

Two sensor arrays were used to characterize flow shadowing. As expected (and seen in 
Figs. \ref{fig:Visualize Shadowing} and \ref{fig:VisFlowShadow}), the downstream whisker exhibits a smaller signal response $||B_{down}||$ when occluded by the upstream whisker. Results in Fig. \ref{fig:VisFlowShadow} also indicate that a more turbulent flow is incident on the downstream whisker. The $||B_{down}||$ signal is more oscillatory and the predicted angle changes frequently. These results were true in both the wind tunnel and under flow from the box fan although the oscillations of both whiskers were greater under the more turbulent flow from the box fan.

\subsubsection{Spacing}
\begin{figure}[h]
\centering
\includegraphics[width=8 cm]{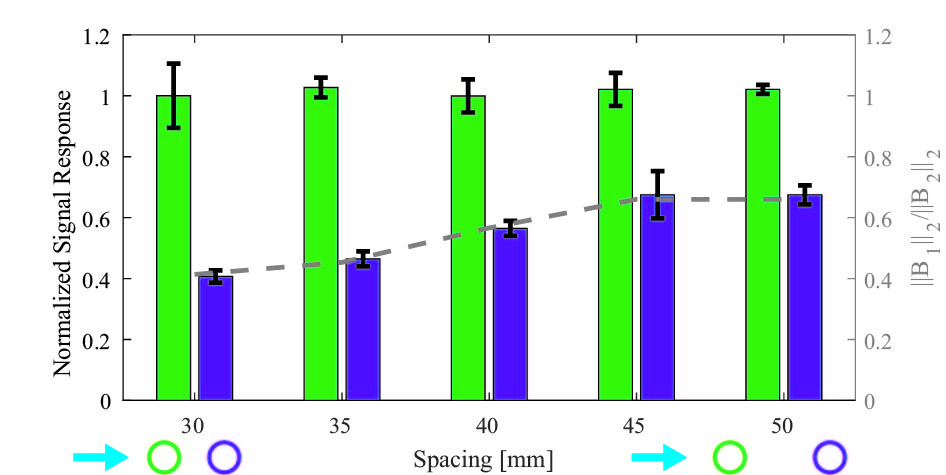}
\caption{The signal response of two whiskers under $100\%$ occlusion were recorded as the spacing, $s$, between the two whiskers was changed.}
\label{fig:spacing}
\end{figure}

 We predicted that the effect of flow shadowing would decrease as the spacing between the whiskers increased. This expectation agreed with the results in Fig. ~\ref{fig:spacing}. The results indicated that a more densely packed array of whiskers will improve the flow shadowing effect. We chose a final spacing of \SI{35}{\milli\meter} to maximize flow shadowing while minimizing the chance of whisker-to-whisker contact.

\subsubsection{Percent Occlusion}
\begin{figure}[h]
\centering
\includegraphics[width=8 cm]{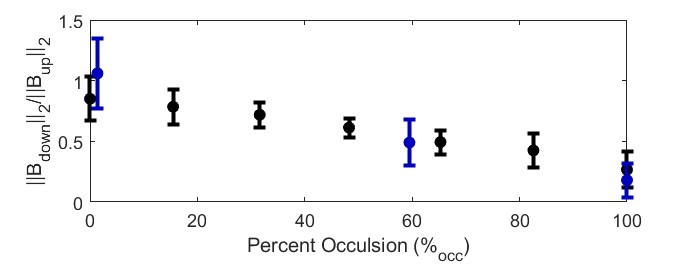}
\caption{Whisker sensors were rotated relative to each other and the heading of flow. This rotation changed $\%_{occ}$ as calculated in Eqn.  \ref{eqn:Oclussion}. When $\%_{occ} \approx 100\%$ the downstream whisker's surface area is fully shadowed by the upstream whisker. When $\%_{occ} \approx 0\%$ the downstream whisker is fully exposed to the flow. The affect of this shadowing on the sensor signal was measured using the fan and wind tunnel setups. The tests were performed with a fan (black) and then repeated in a wind tunnel (blue). The spacing in the fan test was \SI{30}{\milli\meter} and in the wind tunnel \SI{35}{\milli\meter}}
\label{fig:Occluison}
\end{figure}

Fig.~\ref{fig:Occluison}
compares the relative magnitudes of the upstream and downstream whiskers versus the expected percent of occlusion (Eqn. \ref{eqn:Oclussion}). Note that in the results, some flow reaches the downstream whisker even at 100$\%$ occlusion. Flow occlusion was tested in a variety of wind flow types and directions of the whisker; in every result a linear trend was the best descriptor between the percent occlusion and the magnitude ratio. For the algorithm portion of the study, this relationship is represented with the following equation.
\begin{gather}
\frac{||B_{down}||_2}{||B_{up}||_2}=1-0.8*(\%_{occ}/100)
\label{eqn:ExpSigResponse}
\end{gather} 

\begin{figure}[h]
\centering
\includegraphics[width=8 cm]{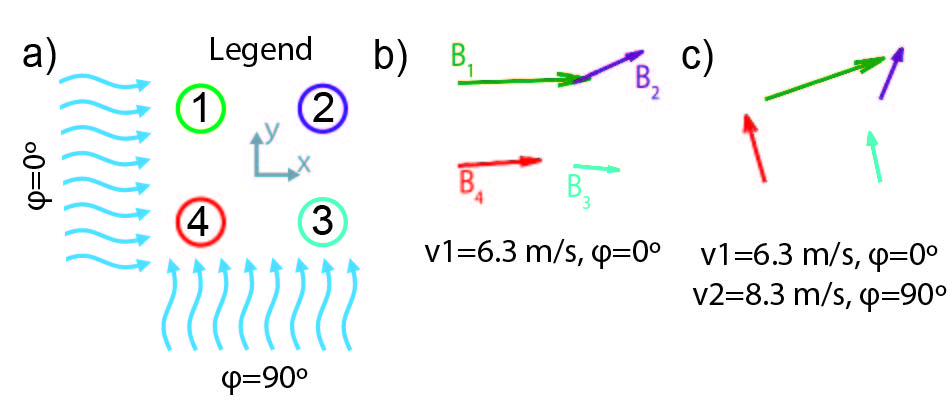}
\caption{Visualization of the flow vectors from the 2x2 array of whiskers in a) under b) one and c) two sources of flow. }
\label{fig:singleFlow}
\end{figure}

\begin{figure*}[bt]
\centering
\includegraphics[width=.87\textwidth]{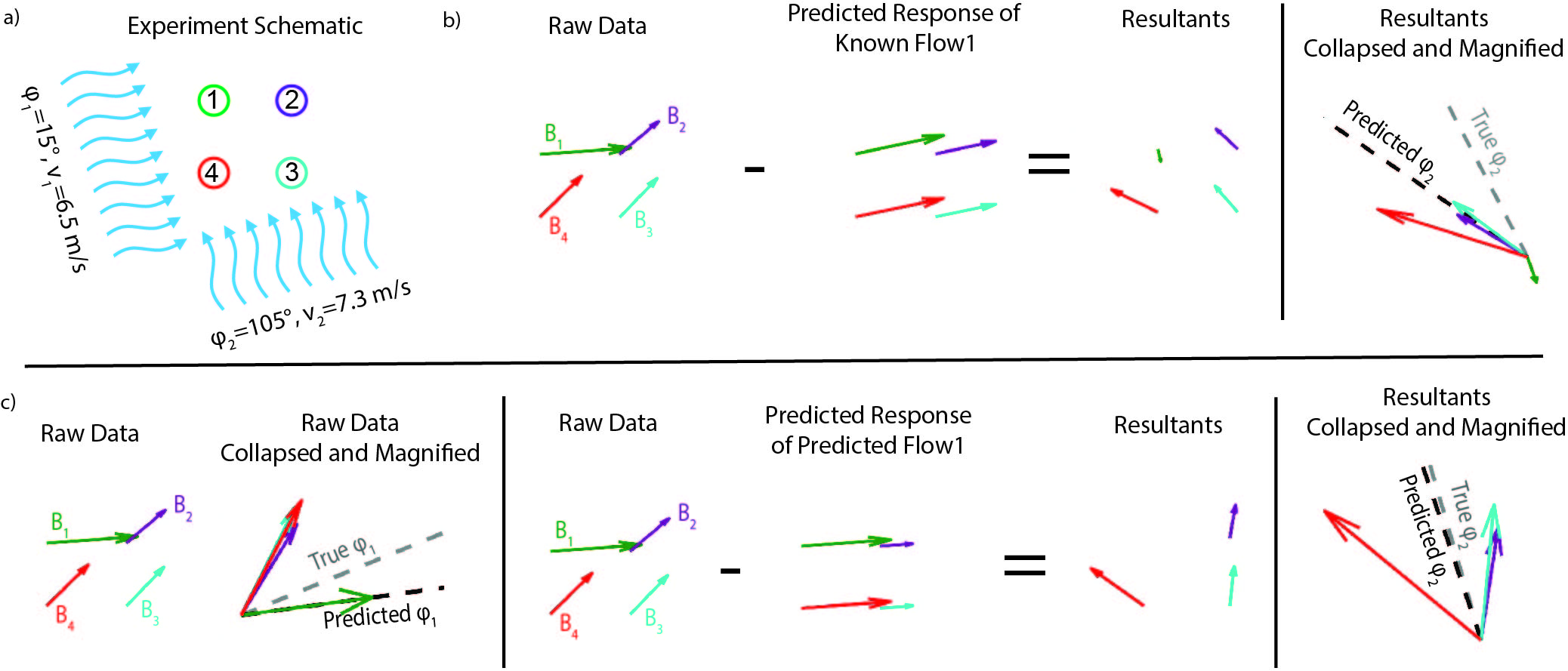}
\caption{a) A single trial of data is used to illustrate how the two algorithms work. b) Algorithm 1 considers a scenario where one of the directions of flow is known ($\varphi_1$) and the algorithm needs to estimate $\hat{\varphi_2}$. c) Algorithm 2 is designed to estimate $\hat{\varphi_1}$ and $\hat{\varphi_2}$ with no prior knowledge. It therefore begins by making a guess about $\hat{\varphi_1}$ before applying a similar method to Algorithm 1. 
\label{fig:Algorithm}}
\end{figure*}

\subsection{Single and Double Flow Characterization}

The effect of occlusion is also seen on a 2x2 array of four whiskers under single flow (Fig. \ref{fig:singleFlow}b). In the example shown, the upstream whiskers' magnitude response $||B||_2$ is larger than the downstream whiskers. The flow direction predicted by the upstream whiskers ($\theta$) is also more accurate to the true flow heading ($\varphi_1 = $ \SI{0}{\degree}). When a second flow source is introduced (Fig. \ref{fig:singleFlow}c), flow blocking becomes multi-directional. In this experiment, whisker 1 is 100 $\%$ occluded from the second flow source at $\varphi_2 = $ \SI{90}{\degree} and therefore the majority of its magnetic field response (B) is due to flow 1 ($\varphi_1 = $ \SI{0}{\degree}). In contrast, whiskers 3 and 4 are $0\%$ occluded from the \SI{90}{\degree} flow which is also a higher velocity flow than flow 1 so their signal is strongly affected by flow 2. Whisker 2 is 100 $\%$ occluded from both sources of flow and its direction prediction is closer to the net flow that we would expect if no flow shadowing occurred. The flow shadowing responses in Fig. \ref{fig:singleFlow}c illustrate the asymmetry in the whisker response across the array that allows us to estimate headings of the two flow sources and prevents the problem shown in (Fig. \ref{fig:DroneNomenclature}c) where all four sensors respond to the average flow.

\subsection{Algorithm}
The algorithms to estimate $\varphi_2$ (if $\varphi_1$ is known) or $\varphi_{1,2}$ (if no flows are known) take advantage of the array's asymmetry. Method 1 uses the known $\varphi_1$, calculates the expected flow across the array for this flow from Eqn. \ref{eqn:ExpSigResponse}, and uses the result to estimate the direction of the second flow (Fig. \ref{fig:Algorithm}b). Method 2 assumes no prior knowledge of the flow headings. A first estimate of $\varphi_1$ is found from the flow direction in the array most separate from the mean (Fig. \ref{fig:Algorithm}c). $\varphi_2$ is then estimated using a similar approach to Method 1. 

Results were collected across at least 2 trials of 24 different combinations of two flows with varying headings and speeds and are summarized in Table \ref{tab:allSummary} for angle combinations with $\alpha = $ \SI{90}{\degree} and \SI{135}{\degree}. For each $\alpha$ and flow speed ratio, a root mean square error (RMSE) was calculated for all trials as
RMSE = $\sqrt{\Sigma_1^n (\varphi_x - \varphi_{x,predicted})^2/n}$.

\begin{table}[h!]
  \begin{center}
    \caption{Flow Heading RMSE Across All Trials}
    \label{tab:allSummary}
    
    \begin{tabular}{| c | c | c | c | c |}
        \hline
            &
            &
            Method 1 &
            \multicolumn{2}{c|}{Method 2} \\
         \hline
          $\alpha$ & 
          $\frac{v1}{v2}$ &
          $\varphi_2$ RMSE & 
          $\varphi_1$ RMSE & 
          $\varphi_2$ RMSE \\
         \hline\hline
          \SI{90}{\degree} &
          0.9 &
          \SI{30.5}{\degree} &
          \SI{9.9}{\degree} &
          \SI{22.9}{\degree} \\
          \SI{90}{\degree} &
          0.7 &
          \SI{16.0}{\degree} &
          \SI{13.2}{\degree} &
          \SI{23.7}{\degree} \\
          \SI{90}{\degree} &
          0.6 &
          \SI{15.4}{\degree} &
          \SI{7.8}{\degree} &
          \SI{7.3}{\degree} \\
          \hline
          \SI{135}{\degree} &
          0.9 &
          \SI{8.9}{\degree} &
          \SI{43.2}{\degree} &
          \SI{26.1}{\degree} \\
          \SI{135}{\degree} &
          0.7 &
          \SI{22.1}{\degree} &
          \SI{36.3}{\degree} &
          \SI{24.9}{\degree} \\
          \SI{135}{\degree} &
          0.6 &
          \SI{20.2}{\degree} &
          \SI{45.0}{\degree} &
          \SI{16.8}{\degree} \\
          
         \hline
    \end{tabular}
  \end{center}
\end{table}

Overall, the calculated RMSE for the various headings across both algorithms were relatively low, especially given that a small array of whiskers (2x2) was used in this study. We did find in test cases when $\alpha = $ \SI{45}{\degree} (not included in Table \ref{tab:allSummary}) that error became significantly higher. In these cases where the two incident flow sources are closer in direction, the small 2x2 array is not able to provide enough asymmetry. The creation of larger arrays to improve accuracy and the smallest distinguishable $\alpha$ is an interesting direction for future study. The velocity ratio also seems to affect the accuracy of the heading predictions although we cannot yet say this with statistical certainty. As expected, prediction estimates seem to improve as the velocity ratio moves further away from 1 (e.g., one flow is significantly faster than the other). Finally, while the results presented are calculated from 50 collected data points, the algorithms will work with as few as 5 data points since these result in a standard deviation below the sensor's RMSE. This will ultimately reduce the time required to estimate the flow vectors.

\section{Conclusion}

This work demonstrates the benefits of flow shadowing, a phenomenon that occurs when whiskers are packed densely in an array, for estimating the headings of multiple flow sources. Using a modified version of whisker-based flow sensors from previous work \cite{Kim2020Whisker-inspired}, we quantified the flow shadowing effect and derived an empirical model. Using a 2x2 whisker array, we showed that flow shadowing creates a direction-dependent asymmetric response. Asymmetry in the array response allowed our algorithms to predict the flow headings with relative accuracy. We expect that more whiskers, a higher density, and improvements to the algorithm can further improve our results, including an estimate of velocity for both flows. The work presented here represents the first steps toward sensors that are capable of a more nuanced understanding of the flow surrounding a variety of unmanned aerial vehicles.

\section*{Acknowledgements}
This research effort was supported by Air Force Office of Scientific Research MURI award number FA9550-19-1-0386. The authors would also like to thank Prof. Satbir Singh and Mr. Ed Wojciechowski for the instruction and guidance in use of the wind tunnel characterization facility.

\bibliographystyle{IEEEtran}
\bibliography{IEEEfull.bib}

\end{document}